\documentclass{article}
\usepackage{arxiv}

\usepackage{subcaption}
\usepackage{graphicx}
\usepackage[utf8]{inputenc} 
\usepackage[T1]{fontenc}    
\usepackage{hyperref}       
\usepackage{url}            
\usepackage{booktabs}       
\usepackage{amsfonts}       
\usepackage{nicefrac}       
\usepackage{microtype}      
\usepackage{amsmath}
\usepackage[english]{babel}

\usepackage{algorithm,algpseudocode}

\title{Mean Field Game GAN}

\author{%
  Shaojun Ma\\
  CSE, Department of Mathematics\\
  Georgia Institute of Technology\\
  Atlanta, GA 30332 \\
  \texttt{shaojunma@gatech.edu} \\
  \And
  Haomin Zhou\\
  Department of Mathematics\\
  Georgia Institute of Technology\\
  Atlanta, GA 30332 \\
  \texttt{hmzhou@math.gatech.edu} \\
  \And
  Hongyuan Zha\\
  School of Data Science\\
  The Chinese University of Hong Kong\\
  Shenzhen, Guangdong 518172\\
  \texttt{zhahy@cuhk.edu.cn} \\
}

\begin{document}

\maketitle

\begin{abstract}
  We propose a novel mean field games (MFGs) based GAN(generative adversarial network) framework. To be specific, we utilize the Hopf formula in density space to rewrite MFGs as a primal-dual problem so that we are able to train the model via neural networks and samples. Our model is flexible due to the freedom of choosing various functionals within the Hopf formula. Moreover, our formulation mathematically avoids Lipschitz-1 constraint. The correctness and efficiency of our method are validated through several experiments.
\end{abstract}

\section{Introduction}
Over the past few years, we have witnessed a great success of generative adversarial networks (GANs) in various applications. Speaking of general settings of GANs, we have two neural networks, one is generator and the other one is discriminator. The goal is to make the generator able to produce samples from noise that matches the distribution of real data. During training process, the discriminator distinguishes the generated sample and real samples while the generator tries to 'fool' the discriminator until an equilibrium state that the discriminator cannot tell any differences between generated samples and real samples.

Wasserstein GAN (WGAN)\cite{wgan} made a breakthrough on computation performance since it adopts a more natural distance metric (Wasserstein distance) other than the Total Variation (TV) distance, the Kullback-Leibler (KL) divergence or the Jensen-Shannon (JS) divergence. However most algorithms related to WGAN is based on the Kantorovich duality form of OT(optimal transport) problem, in which we are facing Lipschitz-1 challenge in training process. Though several approximated methods have been proposed to realize Lipschitz condition in computations, we aim to find a more mathematical method.

MFGs\cite{mfg2,mfg1} are problems that model large populations of interacting agents. They have been widely used in economics, finance and data science. In mean field games, a continuum population of small rational agents play a non-cooperative differential game on a time horizon [0, $T$]. At the optimum, the agents reach a Nash equilibrium (NE), where they can no longer unilaterally improve their objectives. In our case the NE leads to a PDE system consisting of continuity equation and Hamilton-Jacobi equation which represent the dynamical evolutions of the population density and the cost value respectively, following a similar pattern within the framework of Wasserstein distance\cite{apac}. Here the curse of infinite dimensionality in MFGs is overcome by applying Hopf formula in density space.

Utilizing Hopf formula to expand MFGs in density space is introduced in \cite{hjbhopf}. Base on their work, we show that under suitable choice of functionals, the MFGs framework can be reformulated to a Wasserstein-p primal-dual problem. We then handle high dimensional cases by leveraging deep neural networks. Notice that our model is a widely-generalized framework, different choices of Hamiltonian and functional settings lead to different cases. We demonstrate our approach in the context of a series of synthetic and real-world data sets. This paper is organized as following: 1) In Section 2, we review some typical GANs as well as concepts of MFGs. 2) In Section 3, we develop a new GAN formulation by applying a special Lagrangian, our model naturally and mathematically avoids Lipschitz-1 constraint. 3) In Section 4, we empirically demonstrate the effectiveness of our framework through several numerical experiments.

\section{Related Works}

\subsection{Original GAN}

Given a generator network $G_{\theta}$ and a discriminator network $D_{\omega}$, the original GAN objective is to find an mapping that maps a random noise to an expected distribution, it is written as a minimax problem

\begin{align}
    \inf_{G_{\theta}}\sup_{D_{\omega}}\mathbb{E}_{x\sim \rho_{\text{data}}}\big[\log D_{\omega}(x)] + \mathbb{E}_{z\sim \rho_{\text{noise}}}\big[\log(1 - D_{\omega}(G(z))].
\end{align}

The loss function is derived from the cross-entropy between the real and generated distributions. Since we never know the ground truth distribution, when compared with other generative modeling approaches, GAN has lots of computational advantages. In GAN framework, it is easy to do sampling without requiring Markov chain, we utilize expectations which is also easy to compute to construct differentiable loss function. Finally, the framework naturally combine the advantages of neural networks, avoiding curse of dimensionality and is extended to various GANs with different objective functions. The disadvantage of original GAN is that the training process is delicate and unstable due to the reasons theoretically investigated in \cite{arjovsky2017principled}.

\subsection{Wasserstein GAN}
Wasserstein distance provides a better and more natural measure to evaluate the distance between two distributions\cite{villani2003topics}. The objective function of Wasserstein GAN is constructed on the basis of Kantorovich-Rubinstein duality form \cite{villani2008optimal} for distributions $\rho_1$ and $\rho_2$
\begin{align}
    W(\rho_1,\rho_2) = \sup_{\phi, \|\nabla \phi\|\leq 1}\mathbb{E}_{x\sim \rho_1}[\phi(x)] - \mathbb{E}_{x\sim \rho_2}[\phi(x)].
\end{align}
Given duality form, one would like to rewrite the duality form in a GAN framework by setting generator $G_{\theta}$, discriminator $D_{\omega}$ and then minimizing the Wasserstein distance between target distribution and generated distribution
\begin{align}
    \inf_{G_{\theta}}\sup_{D_{\omega}, \|\nabla D_{\omega}\leq 1 \|}\mathbb{E}_{x\sim \rho_{data}}[D_{\omega}(x)] - \mathbb{E}_{z\sim \rho_{\text{noise}}}[D_{\omega}(G_{\theta}(z))].
\end{align}
Though WGAN has made a great success in various applications, it brings a well-known Lipschitz constraint problem, namely, we need to make our discriminator satisfies $\|\nabla D_{\omega}\leq 1 \|$. Lipschitz constraint in WGAN has drawn a lot of attention of researchers, several methods such as weight clipping\cite{wgan}, adding gradient penalty as regularizer\cite{gradientpenalty}, applying weight normalization\cite{weightnorm} and spectral normalization\cite{spenorm}. These methods enable convenience training process, however, besides these approximation-based method, we seek for a new formulation to avoid Lipschitz constraint in WGAN.

\subsection{Other GANs}
A variety of other GANs have been proposed in recent years. These GANs enlarge the application areas and provide great insights of designing generators and discriminators to facilitate the training process. One line of the work is focused on changing objective function such as f-GAN\cite{fgan}, Conditional-GAN\cite{conditionalgan}, LS-GAN\cite{lsgan} and Cycle-GAN\cite{cyclegan}. Another line is focused on developing stable structure of generator and discriminator, including DCGAN\cite{dcgan}, Style-GAN\cite{stylegan} and Big-GAN\cite{biggan}. Though these GANs has made a great success, there is no "best" GAN\cite{bettergan}.

\subsection{Hamilton-Jacobi Equation and Mean Field Games}
Hamilton-Jacobi equation in density space (HJD) plays an important role in optimal control problems\cite{hjd2, hjd3, hjd1}. HJD determines the global information of the system\cite{hjd4, hjd5} and describes the time evolution of the optimal value in density space. In applications, HJD has been shown very effective at modeling population differential games, also known as MFGs which study strategic dynamical interactions in large populations by extending finite players’ differential games\cite{mfg1, mfg2}. Within this context, the agents reach a Nash equilibrium at the optimum, where they can no longer unilaterally improve their objectives. The solutions to these problems are obtained by solving the following partial differential equations (PDEs)\cite{hjbhopf}
\begin{align}
    \begin{cases}
    \partial_t\rho(x,t) + \nabla_x \cdot (\rho(x,t)\nabla_p H(x, \nabla_x\Phi(x,t))) = 0\\
    \partial_t\Phi(x,t) + H(x, \nabla_x\Phi(x,t)) + f(x,\rho(.,t)) = 0\\
    \rho(x,T) = \rho(x), \quad \Phi(x,0) = g(x, \rho(.,0)).
    \end{cases}
\end{align}
where $\rho(x,t)$ represents the population density of individual $x$ at time $t$. $\Phi(x,t)$ represents the velocity of population. We also have each player’s potential energy $f$ and terminal cost $g$. The Hamiltonian $H$ is defined as
\begin{align}
    H(x,p) = \sup_{v}\langle v,p \rangle - L(x,v).
\end{align}
where $L$ can be freely chosen.

Furthermore, a game is called a potential game when there exists a differentiable potential energy $F$ and terminal cost $G$, such that the MFG can be modeled as a optimal control problem in the density space\cite{hjbhopf}
\begin{align}
    U(T,\rho) = \inf_{\rho, v}\{\int_0^T[\int_X L(x,v(x,t))\rho(x,t)dx - F(\rho(.,t))]dt + G(\rho(.,0))\}.
    \label{u1}
\end{align}
where the infimum is taken among all vector fields $v(x, t)$ and densities $\rho(x, t)$ subject to the continuity equation
\begin{align}
\begin{cases}
    \partial_t \rho(x,t) + \nabla \cdot (\rho(x,t)v(x,t)) = 0\quad 0\leq t\leq T\\
    \rho(x,T) = \rho(x).
    \label{con1}
\end{cases}
\end{align}
Our method is derived from this formulation, the details will be introduced in the next section.

\subsection{Optimal Transport}
To better understand the connections between MFGs and WGAN, we also would like to have a review on optimal transport(OT). OT rises as a popular topic in recent years, OT-based theories and formulations have been widely used in fluid mechanics\cite{otfm1}, control\cite{otcontrol1}, GANs\cite{wgan} as well as PDE\cite{learnsde1}. We study OT problem because it provides an optimal map $T$ to pushforward one probability distribution $\mu$ to another distribution $\nu$, with minimum cost $c$. One formulation is Monge’s optimal transport problem\cite{villani2008optimal}
\begin{align}
    \inf_{F}\int_X c(x,F(x)) d\mu (x).
\end{align}

However, the solution of Monge problem may not exist and even if the solution exists, it may not be unique. Another formulation called Kantorovich’s optimal transport problem\cite{benamou1} reads
\begin{align}
\begin{cases}
    \underset{\pi \in \mathcal{P}(X\times Y)}{\inf}\int_{X\times Y} c(x,y) d\pi (x,y)\\
    \int_{Y}d\pi(x,y) = d\mu (x), \quad \int_{X}d\pi(x,y) = d\nu (y).
\end{cases}
\end{align}


Notably, the fluid dynamic formulation of OT\cite{benamou1} reads
\begin{align}
    \begin{cases}
    \underset{v,\rho}{\inf}\int_0^T\int_X \frac{1}{2}\|v(x,t)\|^2\rho(x,t)dxdt\\
    \frac{\partial \rho}{\partial t} + \nabla \cdot (\mu \nu) = 0\\
    \rho(x,0) = \mu, \quad \rho(x,T) = \nu.
    \end{cases}
\end{align}
where $v(x,t)$ is the velocity vector. We solve this problem by applying Lagrange multiplier $\Phi(x,t)$ and optimality conditions, finally we have its dual form
\begin{align}
    \begin{cases}
    \underset{\Phi, \rho}{\sup}\int \Phi(x,T)\rho(x,T) - \int \Phi(x,0)\rho(x,0)\\
    \frac{\partial \Phi}{\partial t} + \frac{1}{2}\|\nabla \Phi\|^2 = 0.
    \end{cases}
    \label{fluidot}
\end{align}
We also refer readers to another related dynamical formulation of OT \cite{li2018constrained}.

\section{Model Derivation}
\subsection{Formulation via Perspective of MFG}
Let's rewrite \ref{u1} and \ref{con1} with defining flux function $m(x,t) = \rho(x,t) v(x,t)$, then we have
\begin{align}
    U(T,\rho) = \inf_{\rho, v}\{\int_0^T[\int_X L(x,\frac{m(x,t)}{\rho(x,t)})\rho(x,t)dx - F(\rho(.,t))]dt + G(\rho(.,0))\},
\end{align}
with
\begin{align}
\begin{cases}
    \partial_t \rho(x,t) + \nabla \cdot m(x,t) = 0 \quad 0\leq t\leq T\\
    \rho(x,T) = \rho(x).
\end{cases}
\end{align}

We solve above primal-dual problem by applying Lagrange multiplier $\Phi(x,s)$
\begin{align}
    U(T,\rho) = \inf_{\rho(x,t), m(x,t)}\sup_{\Phi(x,t)}&\Biggl\{\int_0^T\Biggl[\int_XL\Biggl(x, \frac{m(x,t)}{\rho(x,t)}\Biggr)\rho(x,t)dx - F(\rho(.,t)) \Biggr]dt + G(\rho(.,0))\nonumber\\
    &+ \int_0^T\int_X(\partial_t\rho(x,t) + \nabla \cdot m(x,t))\Phi(x,t)dxdt\Biggr\}.
\end{align}

After integration by parts and reordering it leads to
\begin{align}
    U(T,\rho) =\sup_{\Phi(x,t)}\inf_{\rho(x,t)}&\Biggl\{-\int_0^T\int_X \rho(x,t)H(x,\nabla \Phi(x,t))dxdt - \int_0^T\int_X\rho(x,t)\partial_t\Phi(x,t)dxdt\nonumber\\
    &- \int_0^T F(\rho(.,t))dt + G(\rho(.,0)) - \int_X\rho(x,0)\Phi(x,0)dx \nonumber\\
    &+ \int_X\rho(x,t)\Phi(x,t)dx\Biggr\},
    \label{obj1}
\end{align}

where
\begin{align}
    H(x, \nabla \Phi(x,s)) = \sup_{v}\nabla \Phi\cdot v - L(x,v).
    \label{Hphi}
\end{align}

Now let $L(x,v) = \frac{\|v\|^p}{p}$ where $\|\cdot\|$ can be any norm in Euclidean space.  For example, choose $\|\cdot\| = \|\cdot\|_2$, namely, $\| v \| = \| v \|_2 = \sqrt{v_1^2 + v_2^2 + ... +v_n^2}$. For arbitrary norm $s$, we have the minimizer of \ref{Hphi} and corresponding $H$ as
\begin{align}
    \nabla\Phi &= \|v\|_s^{p-s}\cdot v^{s-1} \\
    H(v) &= \|v\|_s^{p} - \frac{\|v\|_s^p}{p} = \frac{1}{q}\|v\|_s^p.
\end{align}

Furthermore we have
\begin{align}
    \|\nabla \Phi\|_{s'}^q &= (\|v\|_s^{p-s})^q\|v^{s-1}\|_{s'}^q \nonumber\\
    &=[(\int|v|^sdx)^{\frac{p-s}{s}}]^{\frac{p}{p-1}}(\int|v^{s-1}|^{\frac{s}{s-1}}dx)^{\frac{p}{p-1}\frac{s-1}{s}}\nonumber\\
    &= (\int|v|^sdx)^{\frac{1}{s}(p-s)\frac{p}{p-1}+\frac{1}{s}(s-1)\frac{p}{p-1}}\nonumber\\
    &= (\int|v|^sdx)^{\frac{p}{s}}\nonumber\\
    &= \|v\|_s^p\\
    H(\nabla \Phi) &= \frac{1}{q}\|\nabla \Phi\|_{s'}^q \quad
    \text{where} \quad \frac{1}{p}+\frac{1}{q}=1, \quad\frac{1}{s}+\frac{1}{s'} = 1.
\end{align}

Now we are able to plug $H(\nabla \Phi)$ back into \ref{obj1}, let $T=1$ and make $F$ and $G$ as 0, then
\begin{align}
    U(1,\rho) =\sup_{\Phi(x,t)}\inf_{\rho(x,t)}&\Biggl\{\int_0^1\int_X -(\partial_s\Phi(x,t) + \frac{1}{q}\|\nabla \Phi\|_{s'}^q)\rho(x,t)dxdt\nonumber\\
    & - \int_X\rho(x,0)\Phi(x,0)dx + \int_X\rho(x,1)\Phi(x,1)dx\Biggr\}.
    \label{obj2}
\end{align}


Next, we extend \ref{obj2} to a GAN framework. We let $\rho(z,t)$ be the generator that pushforward Gaussian noise distribution to target distribution. Specially, we set $\rho_1$ as the distribution of our ground truth data. We treat $\Phi(x,t)$ as discriminator thus it leads to our objective function

\begin{align}
    L_{\text{MFG-GAN}} = \inf_{\rho(x,t)}\sup_{\Phi(x,t)}\Biggl\{&-\mathbb{E}_{z\sim p(z),t\sim Unif[0,1]}[\partial_t\Phi(\rho(z,t),t)+\frac{1}{q}\|\nabla \Phi(\rho(z,t),t)\|_{s'}^q] \nonumber\\
    &+ \mathbb{E}_{x\sim \rho_1}[\Phi(x,1)] - \mathbb{E}_{z\sim p(z)}[\Phi(\rho(z,0),0)]\Biggr\}.
    \label{obj3}
\end{align}

We solve this minimax problem by algorithm 1. Notably, if we extend \ref{obj3} to a general case with a general Hamiltonian $H$, we revise \ref{obj3} as

\begin{align}
    L = \inf_{\rho(x,t)}\sup_{\Phi(x,t)}\Biggl\{&-\mathbb{E}_{z\sim p(z),t\sim Unif[0,1]}[\partial_t\Phi(G(z,t),t)+ H(x,\Phi, \nabla \Phi)] \nonumber\\
    &+ \mathbb{E}_{x\sim \rho_1}[\Phi(x,1)] - \mathbb{E}_{z\sim p(z)}[\Phi(\rho(z,0),0)]\Biggr\}.
    \label{obj4}
\end{align}

\begin{algorithm}[ht!]
\caption{}
\begin{algorithmic}[1]
\Require Initialize generator $\rho_{\theta}$, discriminator $\Phi_{\omega}$
\For{$\rho_{\theta}$ steps}
\State Sample $x$ from real distribution
\State Sample $z$ from standard normal distribution
\For{$\Phi_{\omega}$ steps}
\State Do gradient ascent on $L_{\text{MFG-GAN}}$ to update $\Phi_{\omega}$
\EndFor
\State Do gradient descent on $L_{\text{MFG-GAN}}$ to update $\rho_{\theta}$
\EndFor
\end{algorithmic}
\end{algorithm}

\subsection{Formulation via Perspective of OT}
Now we are going to derive the same objective function from the perspective of OT. Base on the formulation \ref{fluidot}, we directly apply Lagrange multiplier $\rho(x,s)$ on the constraint to reformulate it as a saddle problem
\begin{align}
    \inf_{\rho}\sup_{\Phi}\Biggl\{\int_0^1\int_X -\Biggl(\frac{\partial \Phi}{\partial t} + \frac{1}{2}\|\nabla \Phi\|^2\Biggr)\rho(x,t)dxdt + \int_X \Phi(x,1)\rho(x,1)dx 
    - \int_X \Phi(x,0)\rho(x,0)dx\Biggr\}.
    \label{fluidot2}
\end{align}

We extend \ref{fluidot2} to a GAN framework: we let $\rho(z,s)$ be the generator and $\Phi(x,s)$ as the discriminator thus finally we achieve the same problem
\begin{align}
    L_{\text{OT-GAN}} = \inf_{\rho(x,t)}\sup_{\Phi(x,t)}\Biggl\{&-\mathbb{E}_{z\sim p(z),t\sim Unif[0,1]}[\partial_t\Phi(\rho(z,t),t)+\frac{1}{2}\|\nabla \Phi(\rho(z,t),t)\|_{}^2] \nonumber\\
    &+ \mathbb{E}_{x\sim \rho_1}[\Phi(x,1)] - \mathbb{E}_{z\sim p(z)}[\Phi(\rho(z,0),0)]\Biggr\},
    \label{obj5}
\end{align}

which has the same structure with \ref{obj3}, thus \ref{obj3} can also be seen as minimizing Wasserstein-2 distance. A similar formulation is introduced in our recent work \cite{highwassgeo}, where we consider two known distributions and compute their Wasserstein geodesic.

\section{Experiments}
Notice that for all experiments we set $s=s'=2$, we adopt fully connected neural networks for both generator and discriminator. In terms of training process, for all synthetic and realistic cases we use the Adam optimizer \cite{kingma2014adam} with learning rate $10^{-4}$.

\begin{figure}[h!]
\begin{subfigure}{.24\textwidth}
  \centering
  \includegraphics[width=1\linewidth]{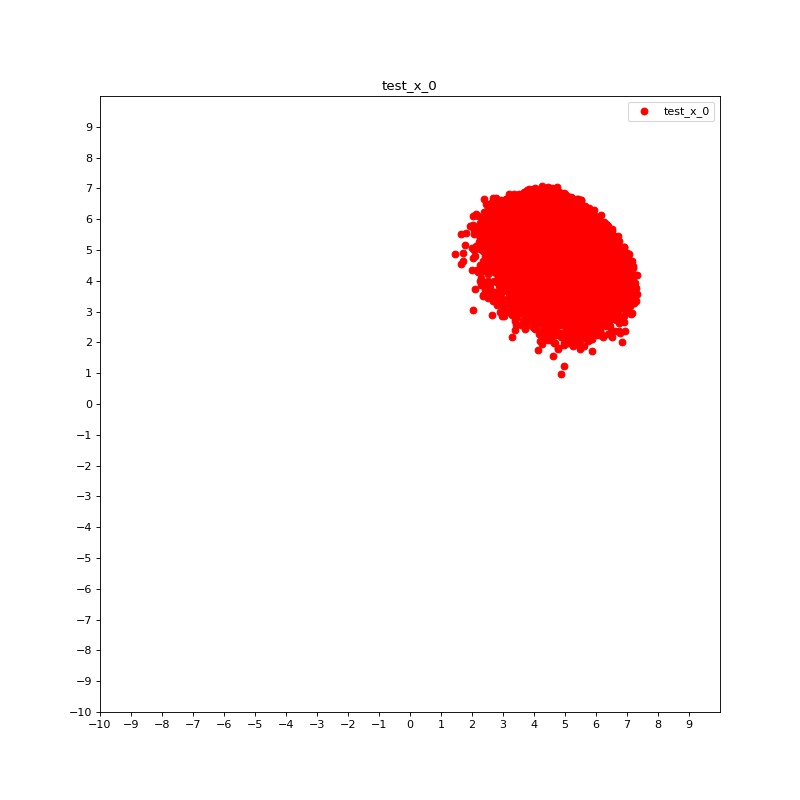}  
  \caption{generated by (z,0)}
\end{subfigure}
\begin{subfigure}{.24\textwidth}
  \centering
  \includegraphics[width=1\linewidth]{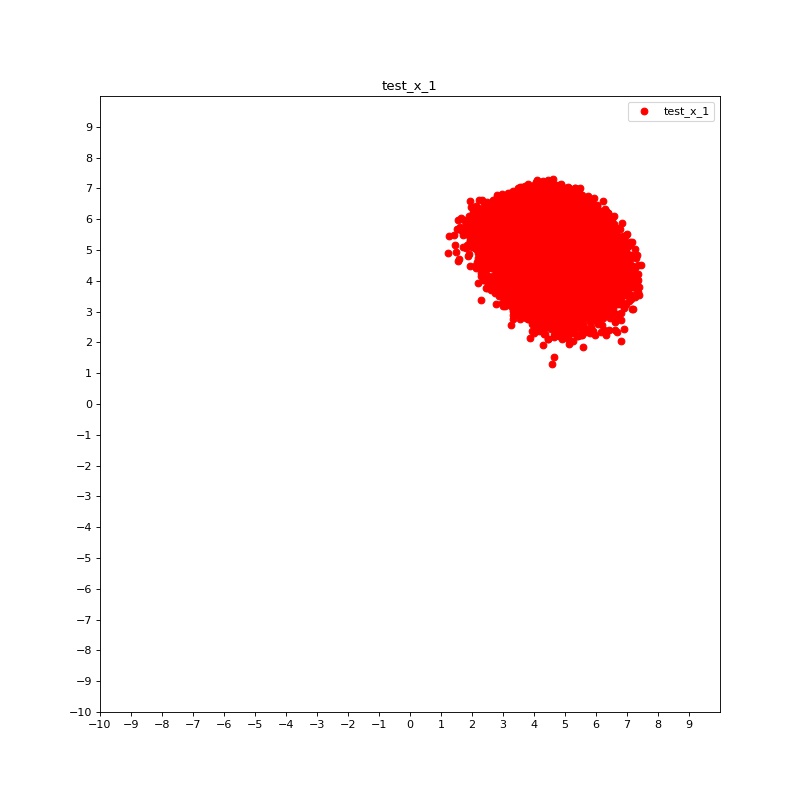}  
  \caption{generated by (z,1)}
\end{subfigure}
\begin{subfigure}{.24\textwidth}
  \centering
  \includegraphics[width=1\linewidth]{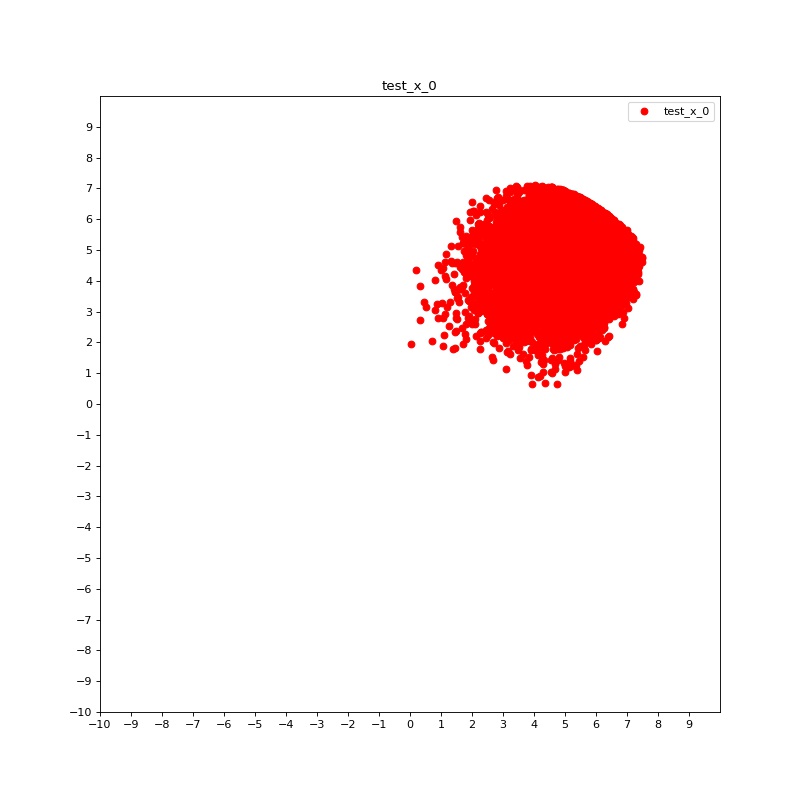}  
  \caption{generated by (z,0)}
\end{subfigure}
\begin{subfigure}{.24\textwidth}
  \centering
  \includegraphics[width=1\linewidth]{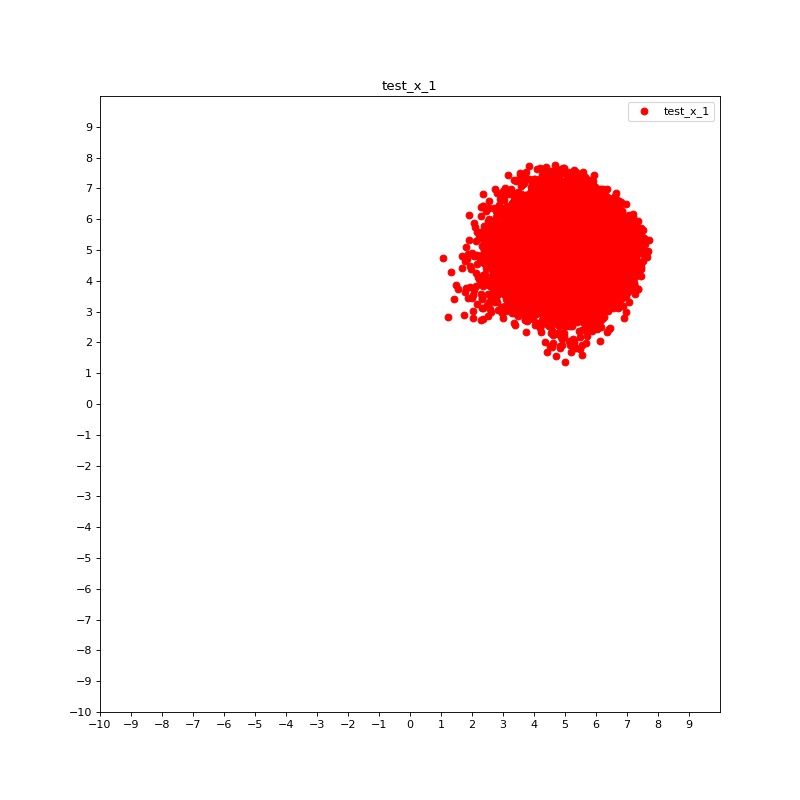}
  \caption{generated by (z,1)}
\end{subfigure}
\caption{Results on 2D synthetic data set}
\label{fig:syn}
\end{figure}

\begin{figure}[h!]
\begin{subfigure}{.24\textwidth}
  \centering
  \includegraphics[width=1\linewidth]{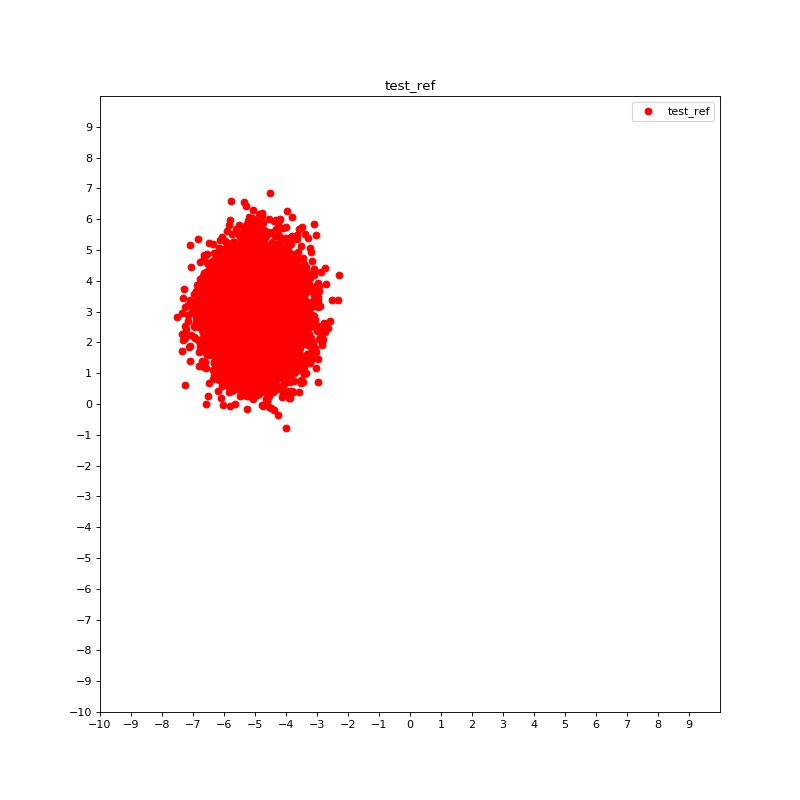}  
  \caption{Ground truth}
\end{subfigure}
\begin{subfigure}{.24\textwidth}
  \centering
  \includegraphics[width=1\linewidth]{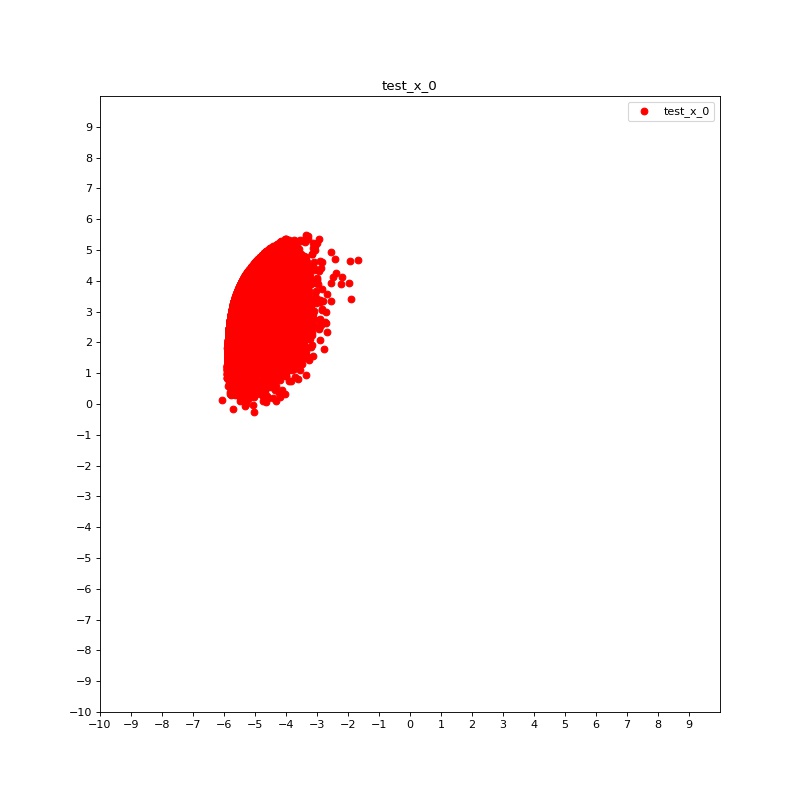}  
  \caption{generated by (z,0)}
\end{subfigure}
\begin{subfigure}{.24\textwidth}
  \centering
  \includegraphics[width=1\linewidth]{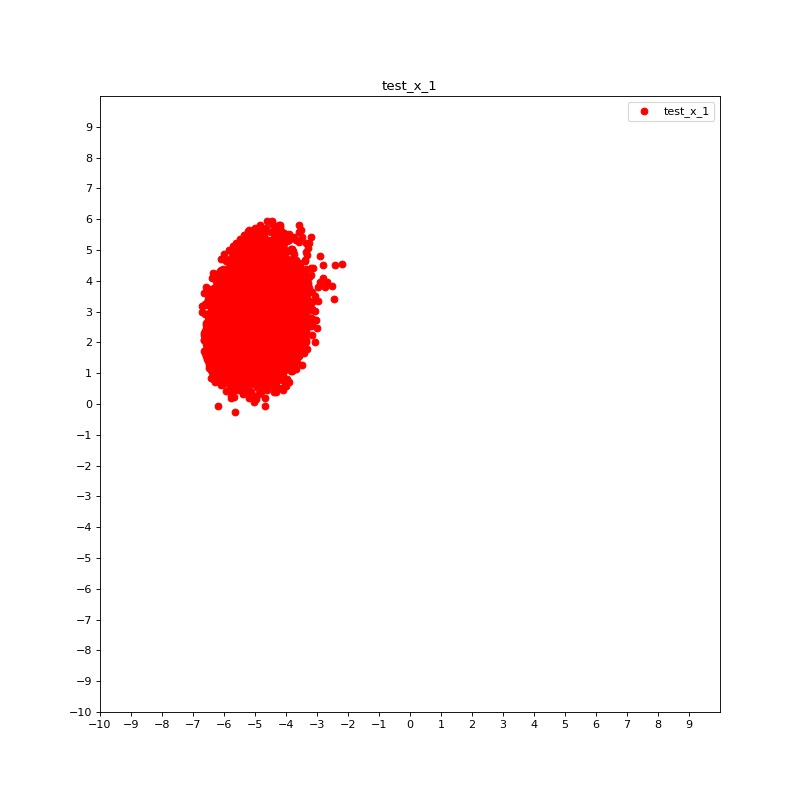}  
  \caption{generated by (z,1)}
\end{subfigure}
\begin{subfigure}{.24\textwidth}
  \centering
  \includegraphics[width=1\linewidth]{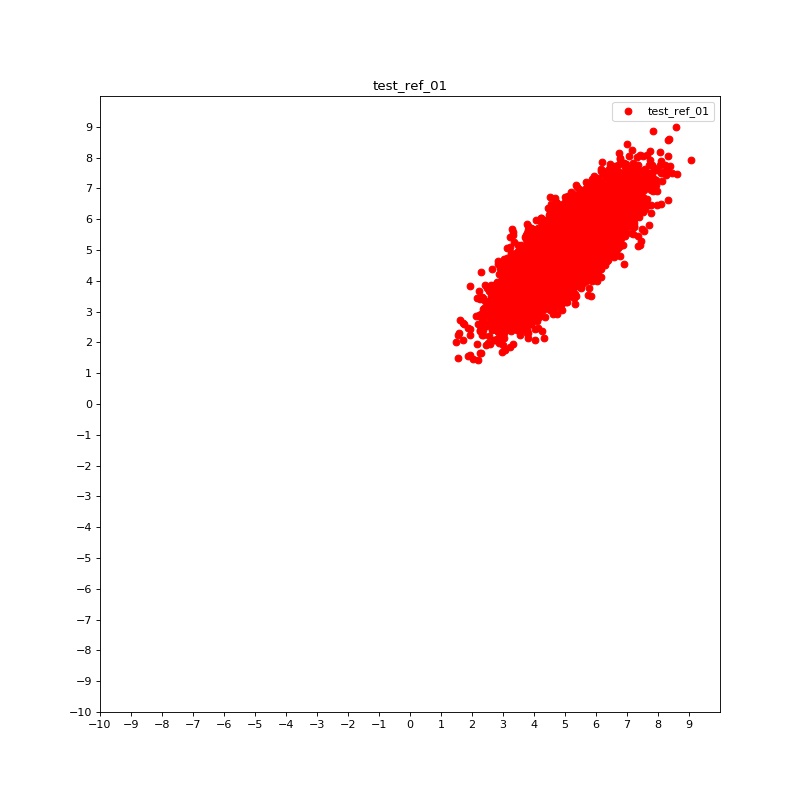}  
  \caption{Ground truth}
\end{subfigure}
\begin{subfigure}{.24\textwidth}
  \centering
  \includegraphics[width=1\linewidth]{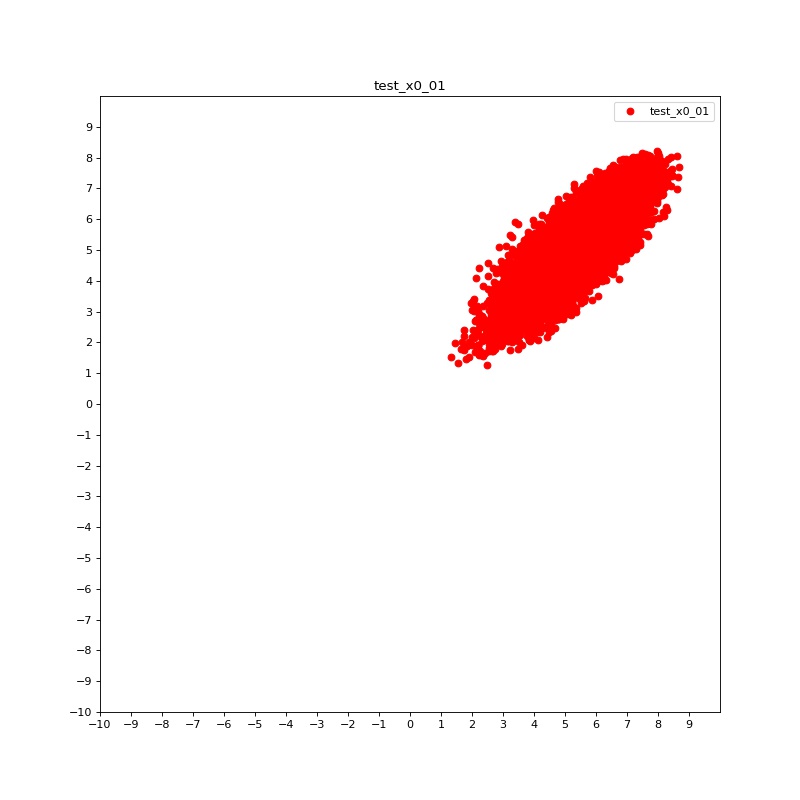}  
  \caption{generated by (z,0)}
\end{subfigure}
\begin{subfigure}{.24\textwidth}
  \centering
  \includegraphics[width=1\linewidth]{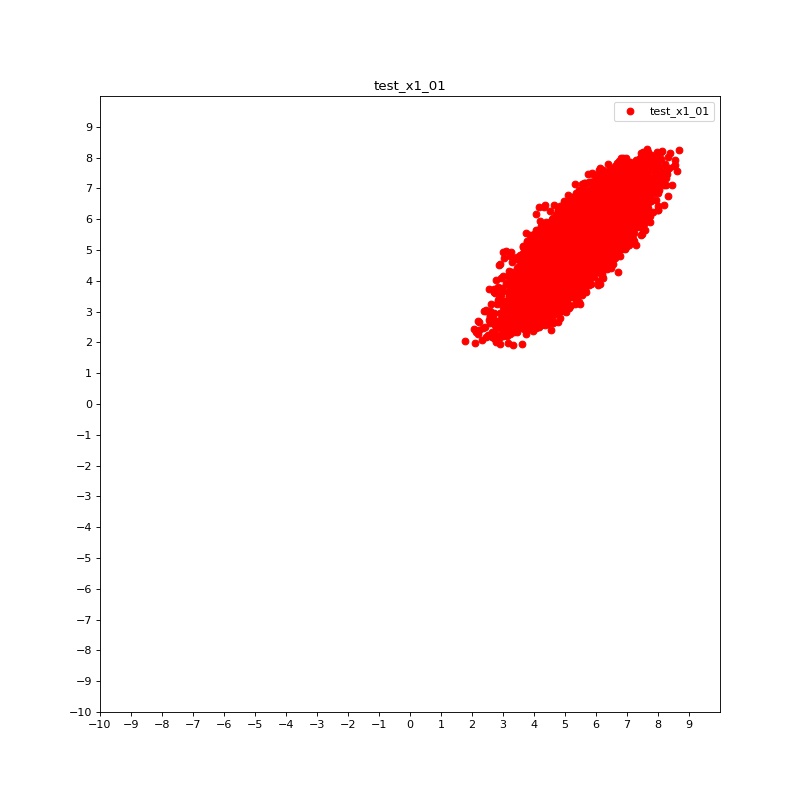}  
  \caption{generated by (z,1)}
  \end{subfigure}
\begin{subfigure}{.24\textwidth}
  \centering
  \includegraphics[width=1\linewidth]{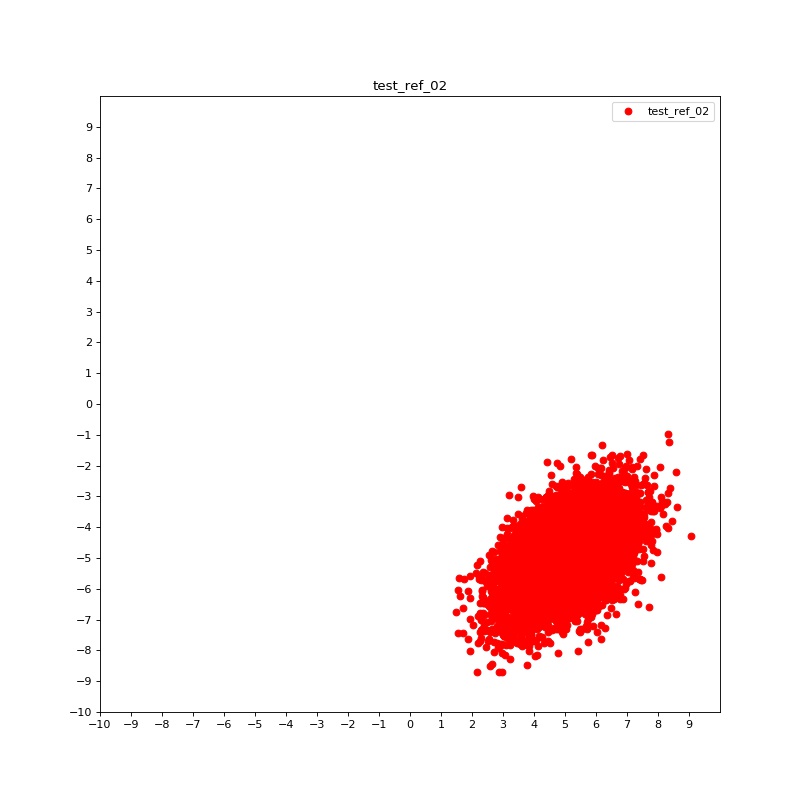}  
  \caption{Ground truth}
\end{subfigure}
\begin{subfigure}{.24\textwidth}
  \centering
  \includegraphics[width=1\linewidth]{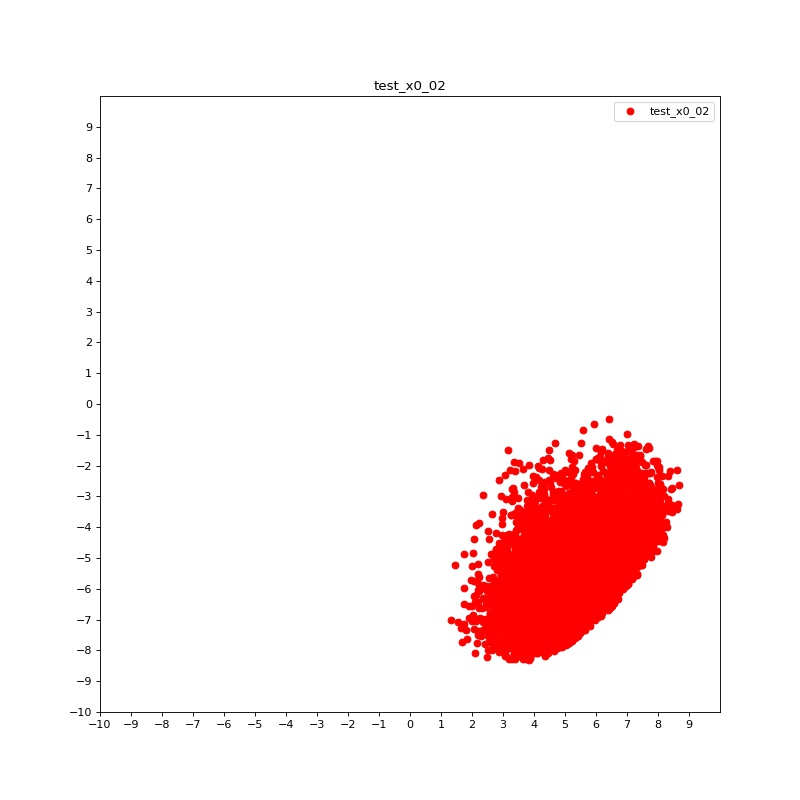}  
  \caption{generated by (z,0)}
\end{subfigure}
\begin{subfigure}{.24\textwidth}
  \centering
  \includegraphics[width=1\linewidth]{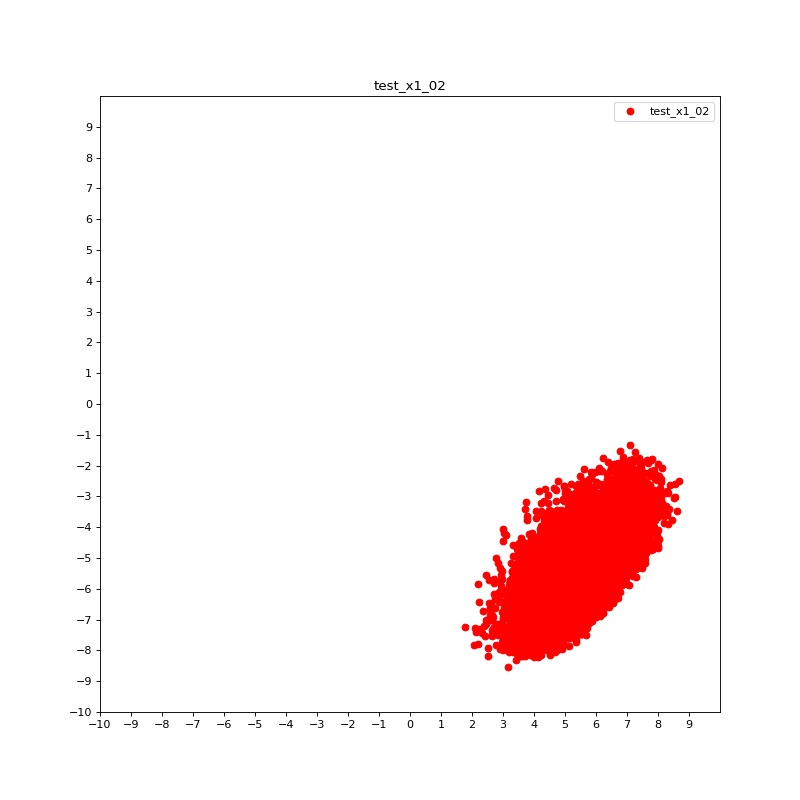}  
  \caption{generated by (z,1)}
\end{subfigure}
\begin{subfigure}{.24\textwidth}
  \centering
  \includegraphics[width=1\linewidth]{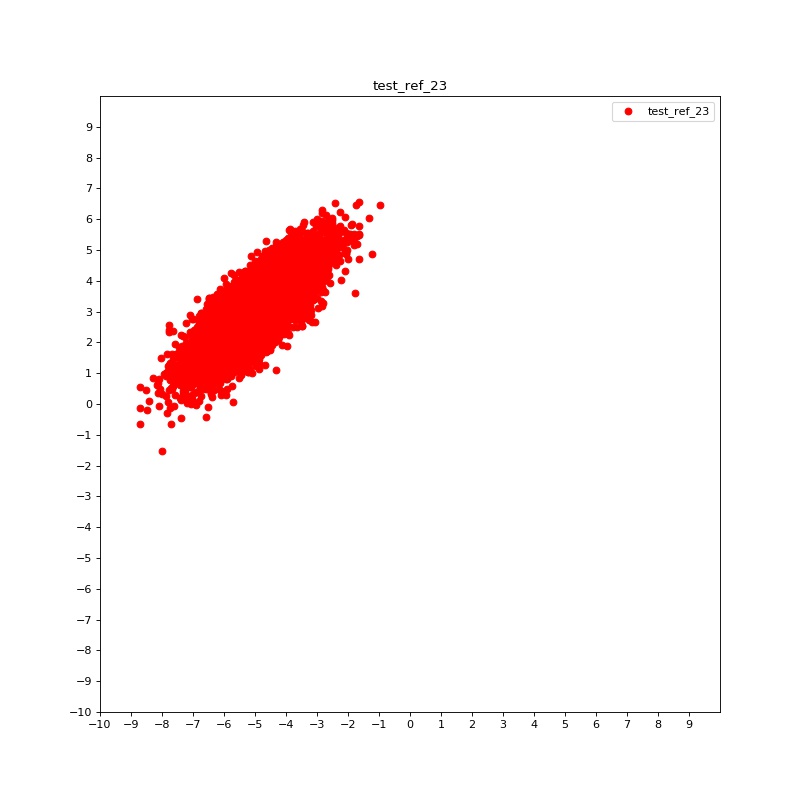}  
  \caption{Ground truth}
  \label{fig:sub-first}
\end{subfigure}
\begin{subfigure}{.24\textwidth}
  \centering
  \includegraphics[width=1\linewidth]{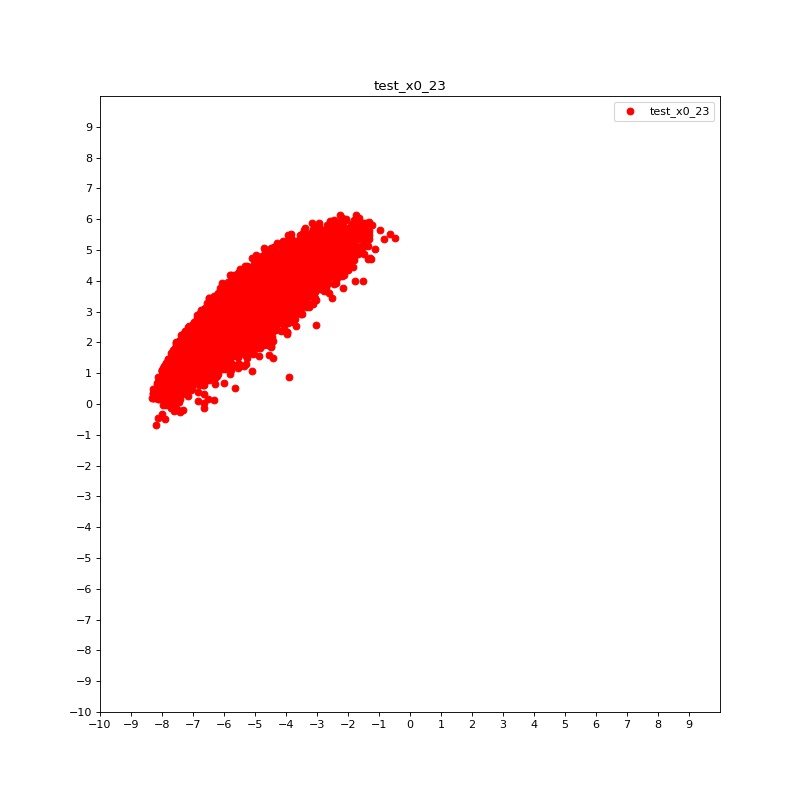}  
  \caption{generated by (z,0)}
  \label{fig:sub-second}
\end{subfigure}
\begin{subfigure}{.24\textwidth}
  \centering
  \includegraphics[width=1\linewidth]{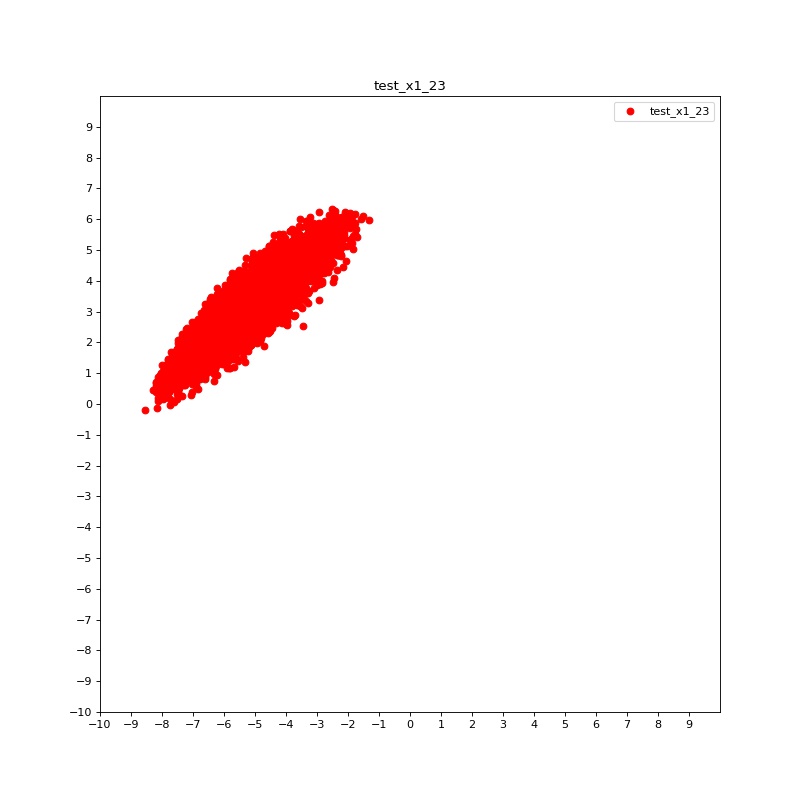}  
  \caption{generated by (z,1)}
  \end{subfigure}
\caption{Results on 10D synthetic data set}
\label{fig:synhigh}
\end{figure}

\subsection{Synthetic}
\textbf{Syn-1}: In this case we test our algorithm on 2D case, the target distribution is a Gaussian $\mathcal{N}(\mu, \Sigma)$, where $\mu$=(5,5) and $\Sigma$= [[1,0],[0,1]], here we choose $q=10, p=1.1$. The generated distributions are shown in Figure \ref{fig:syn} (a) and (b).

\textbf{Syn-2}: Here and in all following tests we set $q=2, p=2$, in this case we still generate a 2D Gaussian distribution, with the same mean and covariance matrix we used in Syn-1. The generated distributions are shown in Figure \ref{fig:syn} (c) and (d).

\textbf{Syn-3}: In this case we generate a 10D Gaussian as our target distribution. Some of 2D projections of the true distributions and generated distributions are shown in Figure \ref{fig:synhigh}.

\begin{figure*}[t!]
  \begin{subfigure}{.3\textwidth}
  \centering
  \includegraphics[width=1\linewidth]{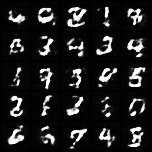}  
  \caption{iteration 18000}
\end{subfigure}\qquad
\begin{subfigure}{.3\textwidth}
  \centering
  \includegraphics[width=1\linewidth]{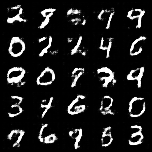}  
  \caption{iteration 52000}
\end{subfigure}\qquad
\begin{subfigure}{.3\textwidth}
  \centering
  \includegraphics[width=1\linewidth]{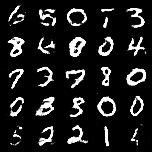}  
  \caption{iteration 146400}
\end{subfigure}
\caption{Generated handwritten digits}
\label{fig:mnist}
\end{figure*}

\subsection{Realistic}
\textbf{MNIST}: We first test our algorithm on MNIST data set. The generated handwritten digits are shown in Figure \ref{fig:mnist}. We believe more carefully designed neural networks such as CNN with extra time input dimension will improve the quality of generated pictures.

\section{Conclusion and Future Directions}
In this paper we derive a new GAN framework based on MFG formulation, by choosing special Hamiltonian and Lagrangian. The same formulation via OT perspective is also presented. The framework avoids Lipschitz-1 constraint and is validated through several synthetic and realistic data sets. We didn't thoroughly investigate the optimal structure of neural networks, we only tried fully connected layers. Some of future works include improving structures of our neural networks, or providing a way to embed time $t$ into CNN.

\bibliography{paper}

\bibliographystyle{plain}
\end{document}